\documentclass[journal]{IEEEtran}
\usepackage{graphicx}
\usepackage{booktabs}
\usepackage{subfigure}
\usepackage{overpic}
\usepackage{amsmath}
\usepackage{flushend}
\usepackage{amssymb}
\usepackage{multirow}
\usepackage{makecell}
\usepackage{fontawesome}
\usepackage{hyperref}
\hypersetup{
	colorlinks=true,
	linkcolor=blue,
	urlcolor=blue,
	citecolor=blue}
\usepackage{colortbl}
\usepackage[dvipsnames]{xcolor}
\definecolor{Bg}{HTML}{e0f1ff}
\usepackage{algorithm, algorithmic}

\begin{document}

\title{Synthetic Aperture Radar Image Change Detection Based on Global Dynamic Context-Aware Network}

\author{
    Baogui Huan, 
    Chuanzheng Gong, 
    Dezhong Chen,
    Feng Gao, \emph{Member, IEEE}, \\
    Junyu Dong, \emph{Member, IEEE}, 
    and Qian Du, \emph{Fellow, IEEE}
\thanks{This work was supported in part by the Natural Science Foundation of Shandong Province under Grant ZR2024MF020, and in part by the Key R \& D Program of Shandong Province under Grant 2025CXPT185. (\textit{Corresponding author: Feng Gao})

Baogui Huan, Chuanzheng Gong, Feng Gao, and Junyu Dong are with the State Key Laboratory of Physical Oceanography, Ocean University of China, Qingdao 266100, China. (email: gaofeng@ouc.edu.cn)

Qian Du is with the Department of Electrical and Computer Engineering, Mississippi State University, Starkville, MS 39762 USA.}}

\markboth{IEEE Journal of Selected Topics in Applied Earth Observations and Remote Sensing}
{Shell}

\maketitle

\begin{abstract}

Convolutional neural networks (CNNs) have been extensively and successfully applied to the task of synthetic aperture radar (SAR) image change detection. However, conventional convolutional layers are inherently limited by their local receptive fields, which mainly capture spatially localized patterns while neglecting the global context that is often crucial for accurately distinguishing subtle or large-scale changes in SAR imagery. Moreover, developing robust and stable detection models in this domain remains particularly challenging due to the scarcity of annotated training samples, which frequently lead to overfitting and unstable convergence during training. To address these limitations, we propose a novel Global Dynamic Context-Aware Network (GDNet) specifically tailored for SAR image change detection. At the core of our approach lies a novel global dynamic convolution module, which adaptively modulates convolution kernel weights according to the global semantic information extracted from the input features. By dynamically incorporating long-range dependencies, this mechanism enables the network to integrate both local detail and global context, thus improving its ability to detect diverse change patterns. In addition, we introduce a carefully designed two-stage Mixup strategy for model training. Unlike conventional single-stage Mixup, our two-stage design generates more diverse and informative training samples, effectively regularizing the model and yielding more stable and reliable classification results even under limited data scenarios. Extensive experiments on three SAR datasets demonstrate the superiority of the proposed GDNet compared to other state-of-the-art methods. These findings highlight the potential of global dynamic modeling and advanced data augmentation strategies for advancing SAR image interpretation. Source codes are available at \url{https://github.com/oucailab/GDNet}.

\end{abstract}

  

\IEEEpeerreviewmaketitle

\section{Introduction}

Remote sensing image change detection refers to the systematic process of comparing multiple images of the same geographic region acquired at different points in time in order to identify, characterize, and analyze changes that have occurred on the Earth’s surface \cite{jiang23tgrs}. These images are most commonly obtained from satellite or aerial-based sensors, which provide consistent and large-scale observations across diverse spatial and temporal resolutions \cite{fs25tgrs}. By highlighting differences between temporal snapshots, change detection techniques allow researchers to monitor a wide range of natural and human-induced phenomena. Typical applications include tracking forest degradation \cite{mathieu22rse}, assessing patterns of urban sprawl \cite{wang22rse}, and rapidly detecting the impacts of natural disasters \cite{ismail22igarss}. Beyond these, change detection is also increasingly used in studying agricultural productivity, wetland dynamics, coastal erosion, and other forms of environmental change \cite{seo23wacv}. Moreover, by capturing the dynamic evolution of Earth systems, remote sensing image change detection enhances our scientific understanding of long-term trends and short-term events, ultimately contributing to the sustainable management of natural resources and improved resilience of human societies \cite{lei23tgrs} \cite{zhu23grsl} \cite{luo23tgrs}.

\begin{figure}[]
\centering
\includegraphics [width=0.7\linewidth]{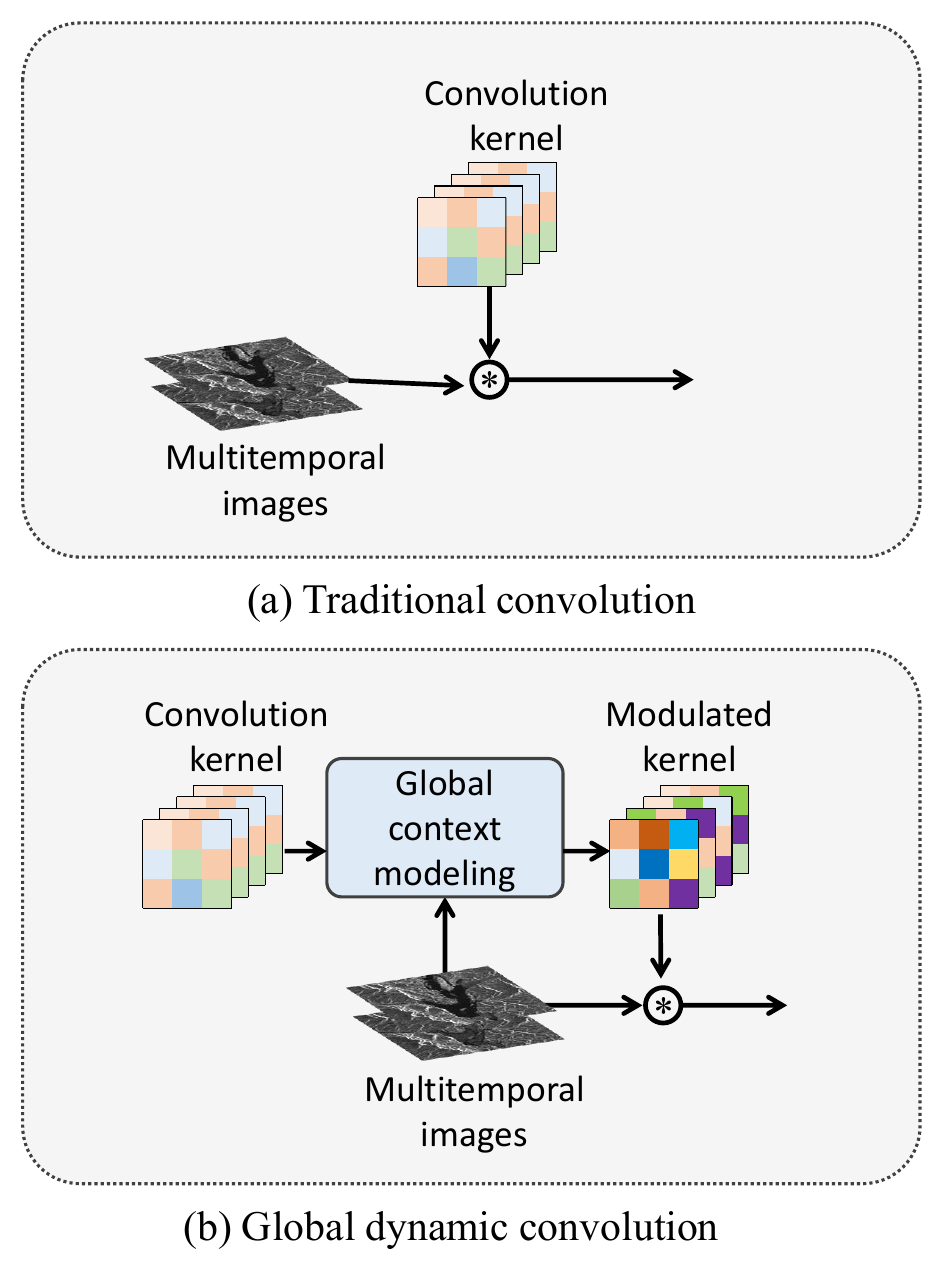}
\caption{Comparison of the traditional convolution with the proposed global dynamic convolution. (a) Traditional convolution commonly uses the same convolutional kernels across different images. Convolutional kernels are content-agnostic and are shared across images. (b) The proposed global dynamic convolution introduces the dynamic mechanism to learn context-aware convolutional kernels. The weights within convolutional kernels are adaptively modified based on the global information of the input features.}
\label{fig_conv_comp}
\end{figure}

In contemporary remote sensing research, a wide variety of sensors are employed for image change detection, including optical, synthetic aperture radar (SAR), hyperspectral, and multispectral sensors \cite{cyq25tgrs}. Among these sensors, SAR is particularly well-suited for remote sensing change detection. It actively transmits its own microwave energy toward the Earth’s surface and records the portion of the signal that is scattered back after interacting with ground targets \cite{liu24jars} \cite{xufang24tgrs} \cite{guoaoqing24jstars}. Unlike passive optical sensors that rely on sunlight, SAR’s active imaging mechanism allows it to operate under virtually all illumination conditions. More importantly, SAR sensors are capable of acquiring images during both day and night and have the unique ability to penetrate atmospheric obstructions such as clouds, haze, and smoke \cite{xu22isprs} \cite{huang15grsl}. This property ensures the consistent acquisition of data across diverse environmental and seasonal conditions, making SAR an invaluable tool for continuous and long-term monitoring. Furthermore, the microwave signals captured by SAR are sensitive to structural and dielectric properties of surface features, enabling the detection of changes in vegetation, soil moisture, urban infrastructure, and water bodies that may be invisible in optical imagery. Owing to these advantages, SAR has become widely recognized as an ideal data source for land-cover monitoring and environmental assessment, and disaster management \cite{wang22dpd} \cite{wang24jars}.

Considerable research efforts have been dedicated to advancing SAR image change detection. The majority of these studies concentrate on self-supervised paradigms, primarily because acquiring accurate and reliable ground-truth labels is often practically infeasible in real-world scenarios. The annotation process is not only labor-intensive and cost-prohibitive but also inherently subject to uncertainty due to temporal inconsistencies, spatial ambiguities, and sensor noise. Consequently, in this paper, we mainly focus on self-supervised SAR image change detection, which offers a more practical and scalable solution for operational applications \cite{li2024causalcd} \cite{jiang2023self}.

In recent years, the emergence of deep learning has significantly transformed the landscape of remote sensing image analysis, owing to its remarkable ability to learn hierarchical and discriminative feature representations directly from data. This paradigm shift has also permeated SAR change detection, where deep neural networks demonstrate strong potential to overcome the limitations of traditional handcrafted feature-based methods. Numerous approaches have been proposed, leveraging architectures such as convolutional neural networks (CNNs) for extracting spatial features \cite{xu22rs}, multi-scale feature modeling to capture contextual information across different resolutions \cite{gao21grsl}, and Transformer-based models to exploit long-range dependencies and global contextual relationships \cite{yanty23tgrs} \cite{zhangxf23tgrs} \cite{xuxt23jstars}. While Transformer-based methods excel at establishing global perspectives, they typically incur excessive computational complexity. Furthermore, they often focus predominantly on global dependencies at the expense of fine-grained local details, which are equally vital for identifying subtle changes and suppressing speckle noise in SAR imagery. Collectively, these advances underscore the growing trend of integrating deep learning into SAR change detection and highlight the need for further exploration of robust, self-supervised frameworks that can efficiently balance global context and local feature extraction.

Although deep learning-based change detection methods have achieved remarkable performance in recent years, there still exist several critical limitations: \textbf{(1) \textit{Global context and local feature dynamic aggregation}.} Traditional CNN-based methods rely heavily on convolutional layers, which are inherently limited to capturing local spatial information within a restricted receptive field. As a result, the broader contextual dependencies across the entire image are often ignored or only weakly modeled. In the case of SAR change detection, where changes can occur across large spatial extents or involve subtle variations influenced by surrounding structures, it becomes particularly important to dynamically integrate both local fine-grained details and global contextual cues. Without such integration, the change detection framework may fail to distinguish meaningful changes from irrelevant noise or background variations. \textbf{(2) \textit{Instability of training}.} Another major challenge lies in the instability of training deep models for SAR change detection. The task itself is constrained by the scarcity of high-quality, well-annotated training samples, since manual labeling in SAR images is costly and time-consuming. To mitigate this limitation, researchers often adopt data augmentation techniques such as Mixup \cite{zhang17mixup} and CutMix \cite{YunHCOYC19Cutmix}, which artificially expand the diversity of the training data and enrich the feature representation space. While these strategies are effective in improving generalization, they may also introduce inconsistent label assignments. The inconsistent label assignments lead to unstable training dynamics and degraded performance. Therefore, developing mechanisms that can retain the benefits of data augmentation while suppressing its negative impacts remains a crucial task for advancing robust SAR change detection performance.

To address the aforementioned two challenges, we propose a novel \textbf{G}lobal \textbf{D}ynamic Context-aware \textbf{N}etwork (\textbf{GDNet}) for  change detection in multitemporal SAR images. The core idea of GDNet is to simultaneously strengthen the aggregation between global context and local features while stabilizing the training process under data augmentation strategies. First, to effectively capture the long-range dependencies that are often overlooked by traditional convolutional neural networks, we design a global dynamic convolution module. Unlike conventional convolution layers with fixed kernel weights, this module adaptively adjusts the convolutional weights according to the global semantic information derived from the input features. In this way, the local feature extraction process is dynamically guided by the global context, allowing the model to better discriminate subtle changes from background noise or irrelevant variations, as shown in Fig. \ref{fig_conv_comp}. Second, to alleviate the instability caused by limited labeled SAR samples and the side effects of aggressive data augmentation, we introduce a two-stage Mixup training strategy. This strategy not only preserves the benefits of enriched feature diversity but also reduces the adverse impact of inconsistent label assignments, leading to a more stable and reliable optimization process. Extensive experiments conducted on three publicly available SAR datasets verify the effectiveness of our design, where GDNet consistently outperforms several state-of-the-art approaches in terms of both accuracy and robustness, demonstrating its strong potential for practical SAR change detection applications.

In summary, the contributions of this paper are three-fold:

\begin{itemize}
	
\item \textbf{Dynamic integration of global features into convolution layers.} By jointly exploiting both spatial and channel-wise feature aggregations, our approach is able to generate content-adaptive convolution kernels rather than relying on fixed weights. It enhances the model’s capacity to capture subtle but meaningful changes.

\item \textbf{Two-stage Mixup for stable training.} We developed a novel two-stage Mixup scheme tailored for SAR change detection. This approach not only enriches the training data distribution to improve generalization but also carefully balances the label mixing process across different stages. 

\item \textbf{Extensive evaluation and open-source contribution.} Comprehensive experiments conducted on three benchmark SAR datasets validate the superiority of the proposed GDNet. Furthermore, to encourage reproducibility and facilitate further research in the community, we will release the codes and data.

\end{itemize}

\section{Related Work}

\subsection{Deep Learning-Based SAR Image Change Detection}

Many CNN-based methods have been proposed to exploit the fine-grained feature representations of multitemporal images. Gao et al. \cite{gaoyh19jstars} proposed a simple yet effective channel weighting-based CNN for SAR change detection. A pooling operator is used to aggregate channel-wise information, and meaningful channel features are emphasized. Ma et al. \cite{malr23icip} presented a CNN-based method which explores the spatial and frequency domain features of SAR images in parallel to improve detection performance. Kaneko et al. \cite{kaneko21igarss} proposed an inter-orbit change detection network, which incorporates the satellite orbit information into the convolutional network. Luppino et al. \cite{luppino22tgrs} designed two deep CNNs for heterogeneous image change detection. An affinity-based change prior is learned from the input data to drive the training process. Hou et al. \cite{houb20tgrs} introduced an end-to-end dual-branch convolutional network. Instead of employing the conventional image domain for differencing, the network conducts differencing in the feature domain. It significantly mitigates the loss of valuable information for change pixel discrimination. Gao et al. \cite{gao19cwnn} introduced a convolutional-wavelet neural network for SAR change detection. They incorporated the dual-tree complex wavelet transform into convolutional neural networks to classify pixels as changed or unchanged. This integration is proved to be effective in reducing the impact of speckle noise. Li et al. \cite{lixh22grsl} proposed a multi-scale fully convolutional neural network for remote sensing image change detection, which used multi-scale convolution kernels to extract the detailed features of the ground object features.

Some Transformer-based methods are recently proposed for SAR change detection. Xie et al. \cite{xjw24grsl} proposed wavelet-based self-attention block that leverages wavelet transform to downsample key and value components without information loss, thereby preserving high-frequency details while expanding the receptive field. Dong et al. \cite{dhh24jstars} employed context-augmented Swin Transformer with global channel-wise aggregation to jointly model local, global, and multi-scale context information for SAR change detection. Li et al. \cite{lll24rs} combined multi-directional feature learning with a multi-scale self-attention mechanism to jointly exploit local edge cues and global context for SAR change detection. Zhong et al. \cite{zwk25grsl} employed a dual-branch backbone with multi-directional convolution and Transformer block. Gated cross-fusion further integrates multi-dimensional features with cross attention, effectively highlighting change-related information. To achieve a balance between local feature extraction and global attention, Zhang et al. \cite{zhang2023convolution} proposed a Convolution and Attention Mixer (CAMixer), which utilizes a parallel design of shift convolution and self-attention, complemented by a gated feed-forward network to suppress speckle noise. To address the challenge of heterogeneous feature representation between different modalities (e.g., optical and SAR), Jing et al. \cite{jing2025hetecd} designed HeteCD, employing a Siamese Transformer encoder and a feature consistency alignment strategy to bridge the modality gap. Furthermore, to mitigate the severe interference of speckle noise and the issue of limited samples in polarimetric SAR (PolSAR) images, Xu et al. \cite{xu2025statistic} introduced a Statistic-Guided Difference Enhancement Graph Transformer (SDEGT) to mine spatial correlations via a superpixel-level graph. Similarly, to tackle the difficulty of synchronously capturing fine-grained local details and long-range dependencies, Xu et al. \cite{xu2024dual} proposed a Dual Attention-Based Global-Local Feature Extraction Network (DA-GLN), which combines a deep residual shrinkage network with a pooling-based Vision Transformer (PiT).

Despite their strong global modeling capabilities, a common limitation of these advanced methods is their heavy reliance on computationally intensive self-attention mechanisms, complex graph constructions, or elaborate dual-branch fusion strategies. Consequently, they often incur excessive computational overhead and require massive memory consumption. Moreover, under scenarios with limited training samples, such highly parameterized models are prone to unstable optimization and may struggle to efficiently preserve the fine-grained local structural details crucial for SAR change detection. On the other hand, traditional CNN-based change detection methods are computationally efficient, but they commonly use the same fixed convolutional kernels across different images. Specifically, standard convolutional kernels are content-agnostic and are shared across spatial locations, which fundamentally limits their capacity to adaptively model global contextual variations.

To break this limitation and achieve a better trade-off between global context modeling and computational efficiency, content-adaptive convolutional kernels have been recently proposed for enhanced feature learning, such as CondConv \cite{condconv} and DynamicConv \cite{DynamicConv}. These methods predict coefficients for several experts, and the convolution kernels are dynamically modified based on the combination of these experts. However, most existing SAR change detection methods neglect the use of dynamic kernels. Different from current related works, our model introduces a dynamic mechanism to learn context-aware convolutional kernels that explicitly aggregate global information, thereby bringing a global perspective into highly efficient local convolutional operations for SAR change detection. By doing so, the network adaptively suppresses inherent speckle noise while highlighting subtle changes. 

\begin{figure*}[t]
\centering
\includegraphics [width=0.8\linewidth]{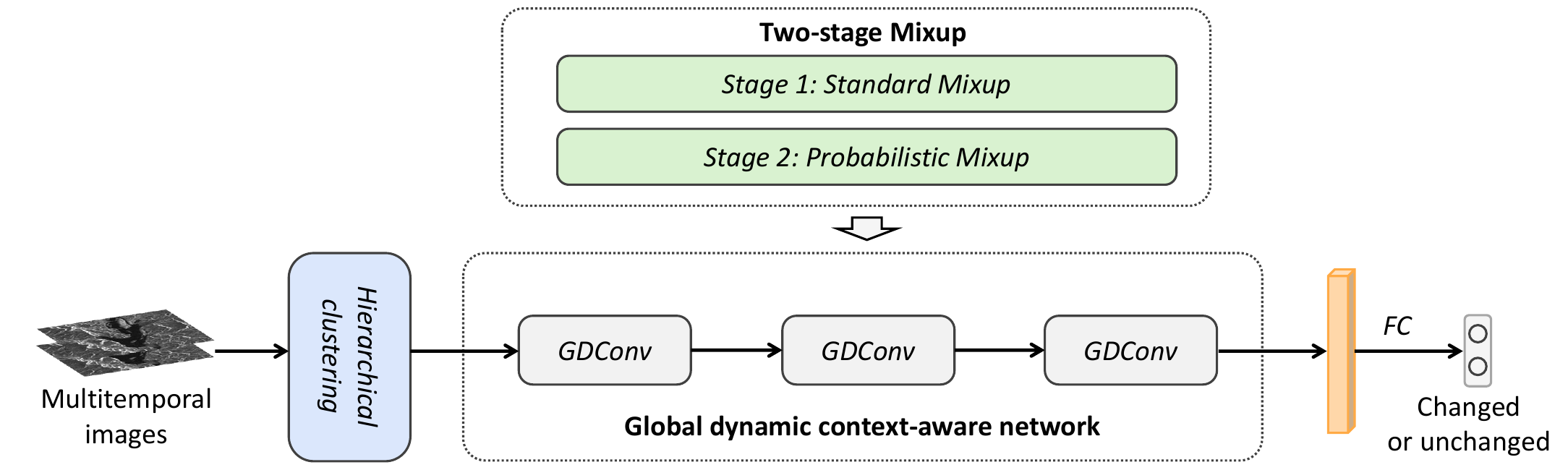}
\caption{Schematic illustration of the change detection method based on global dynamic context-aware network (GDNet).}
\label{fig_frame}
\end{figure*}

\subsection{Data Augmentation for Remote Sensing Applications}

Deep learning-based remote sensing image interpretation methods commonly suffer from the problem of small training data. Basic image manipulations (such as rotations, flips, and noise injections) are widely used for data augmentation \cite{mi25jstars}. Chen et al. \cite{chen22tgrs} investigated the synthesis of data to improve change detection performance. They devised instance-level change augmentation, utilizing generative adversarial training to generate bitemporal images featuring diverse and abundant building changes. Haut et al. \cite{haut19grsl} introduced a random occlusion-based data augmentation method for hyperspectral image classification. This method uses randomly occluded pixels in distinct rectangular spatial regions, thereby creating training images with varying levels of occlusion. This method helps mitigate the risk of overfitting. Xu et al. \cite{xkj20grsl} utilized rotation, flipping, and cropping techniques to enhance the generalization ability of a multi-layer feature fusion network, thereby improving the performance of remote sensing scene classification. Wang et al. \cite{wang23tnnls} employed data augmentation to learn more transferable and diverse feature expressions. Therefore, the data augmentation effectively reduces the representation bias in feature extraction, and therefore improves the multi-modal remote sensing image classification performance. Tang et al. \cite{tang25cvpr} proposed layout-controllable diffusion model supports both horizontal and rotated bounding box–conditioned generation, enabling the creation of high-quality synthetic remote sensing images with controllable layouts and object categories.

Some methods attempt to mix images for data augmentation. For example, Mixup \cite{zhang17mixup} generates weighted combinations of random image pairs from the training data. CutMix \cite{YunHCOYC19Cutmix} cuts one image patch and pastes it to another training image. Specifically, during training, Mixup generates virtual feature-level vectors as follows:
\begin{equation}
    \tilde{x} = \lambda x_i +(1-\lambda)x_j,
\end{equation}
\begin{equation}
    \tilde{y} = \lambda y_i +(1-\lambda)y_j,
\end{equation}
where $(x_i, y_i)$ and $(x_j, y_j)$ are two feature-label vectors randomly selected from the training data in a mini-batch. $\lambda\in[0, 1]$ and it is sampled from Beta distribution.

Although Mixup is widely recognized for its ability to significantly enrich the sample representation space and improve the generalization ability of deep neural networks, it inevitably introduces certain drawbacks. Specifically, by interpolating both features and labels, Mixup can generate uncertain or ambiguous training samples that do not always correspond to realistic data distributions. This ambiguity often leads to unstable optimization dynamics, causing the network to oscillate or converge more slowly during training. To mitigate this issue, we propose a two-stage Mixup strategy. In the first stage, Mixup is applied as usual to expand the representation diversity and encourage the model to learn smoother decision boundaries. In the second stage, the mixing operation is applied probabilistically with a linearly decaying rate, gradually shifting the model to train more frequently on unaltered, original samples. This staged design retains the representation-enriching benefits of Mixup in the early phase while ensuring stable optimization and robust convergence in the later phase, ultimately leading to more reliable convergence and improved performance in SAR change detection tasks.

\section{Methodology}

\subsection{Overview of the Proposed GDNet}

Existing CNN-based change detection methods rely on a sliding window convolution mechanism to capture local features, leaving the global context insufficiently explored. While self-attention mechanisms are widely used for global context modeling, they typically incur high computational complexity and are relatively weak in extracting fine-grained local details, which are crucial for SAR image change detection. To overcome these limitations, we design a global dynamic convolution (GDConv) for context-aware feature modeling. Compared to self-attention, GDConv effectively integrates global semantic cues with local structural information by dynamically modulating convolutional kernels, thereby significantly enhancing the capability to represent detailed features while maintaining a lower computational footprint.

Given two multitemporal coregistered SAR images, denoted as $I_1$ and $I_2$, which are acquired over the same geographical region, the objective of our method is to compute an accurate change map that highlights regions of significant surface variation. The overall framework of the proposed GDNet is illustrated in Fig. \ref{fig_frame}. First, a preclassification step is performed to automatically select training samples with a high likelihood of belonging to either the ``changed" or ``unchanged" category. This step reduces noise in the training data and ensures that the subsequent learning process focuses on more reliable examples. Second, three global dynamic convolution layers are introduced to jointly capture both global contextual dependencies and local spatial details. By dynamically adjusting the convolutional kernels according to the global information of the input, these layers effectively enhance the feature representations of potential change regions. Third, to stabilize the training process and mitigate the adverse effects introduced by data augmentation, we employ an easy-to-use two-stage Mixup regularization strategy. This design allows the model to benefit from enriched sample diversity while avoiding instability in the later stage of training. Finally, the extracted multiscale and globally enhanced features are reshaped into a feature vector and passed through a fully connected (FC) layer. The FC layer performs the final binary classification, categorizing each pixel into either “changed” or “unchanged,” thereby generating the final change detection map.

For the preclassification stage, we adopt the hierarchical fuzzy c-means (FCM) clustering method \cite{gao16grsl}, which has demonstrated strong capability in partitioning difference images into semantically meaningful regions. This preclassification process generates pseudo-labels directly from the inherent structure of the input data without requiring any human-annotated ground truth. Therefore, the proposed GDNet essentially operates under a self-supervised learning paradigm, where the model learns from the autonomously generated supervisory signals. Specifically, the constructed difference image is divided into three distinct clusters by FCM: the changed class, the unchanged class, and the intermediate class. Among these, pixels assigned to the changed class are treated as positive samples, while those in the unchanged class serve as negative samples for training the proposed GDNet. Pixels in the intermediate class, which are characterized by high uncertainty and potential ambiguity, are excluded to avoid introducing noisy labels that could compromise the reliability of the training process.

Since our method performs change detection in a pixel-wise manner rather than full-image segmentation, we extract small local image patches of size $r \times r$ centered on each given pixel from $I_1, I_2$, and the corresponding difference image, respectively. For the proposed network, this $r \times r$ patch constitutes a complete independent input sample. These three patches, containing complementary temporal and differential information, are then concatenated along the channel dimension to form a new image patch of size $3 \times r \times r$. The newly constructed patches are subsequently passed through three carefully designed global dynamic convolution layers. These layers jointly exploit global contextual cues and local spatial structures, and thereby enhance discriminability between changed and unchanged regions.

In the following subsections, we provide a detailed description of the proposed global dynamic convolution mechanism as well as the two-stage Mixup strategy.

\subsection{Global Dynamic Convolution}

\begin{figure}[]
\centering
\includegraphics [width=\linewidth]{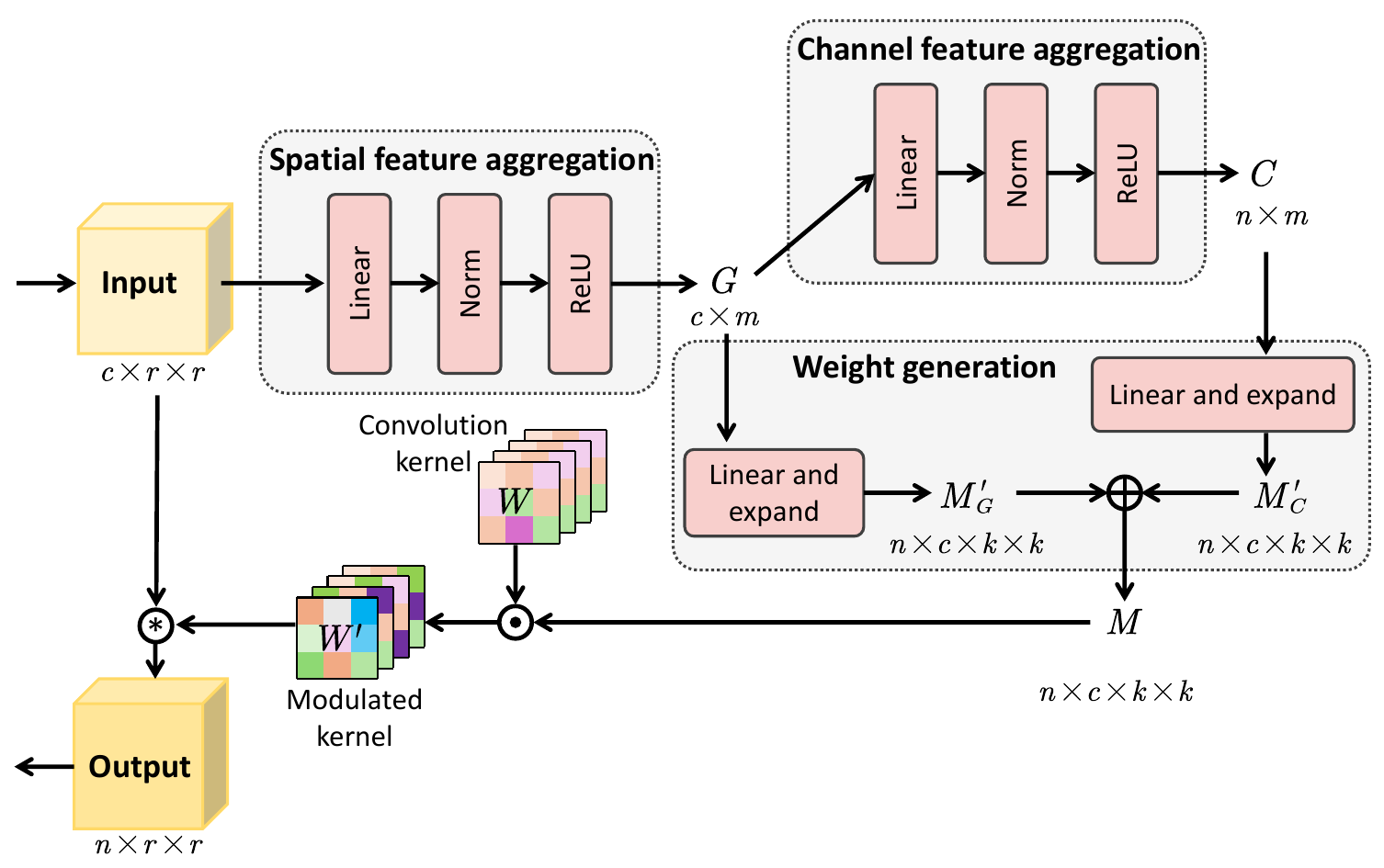}
\caption{Illustration of the global dynamic convolution (GDConv). Firstly, spatial aggregation features $G$ are encoded by linear embedding, and then these features are projected to $C$ by channel feature aggregation module. Finally, $G$ and $C$ are combined to generate weight for the convolution kernel modulation. Convolution is performed by using the new convolution kernel.}
\label{fig_gdc} 
\end{figure}

We introduce a global dynamic convolution (GDConv) to extract context-aware features for SAR change detection. It should be clarified that since our network takes $r \times r$ patches as independent input samples, the term "global" here accurately refers to the global context within the input patch (i.e., sample-level global information). Then the convolution kernel is modulated by an effective combination of global context and local features. Details of the GDConv are shown in Fig. \ref{fig_gdc}. Specifically, GDConv consists of three parts: spatial feature aggregation, channel feature aggregation, and weight generation. Note that the input feature is a feature map of size $c\times r \times r$, and the output feature is a feature map of size $n \times r \times r$. Here $c$ is the number of input channels, and $n$ is the number of output channels. The kernel in each convolution is $W\in\mathbb{R}^{n\times c\times k\times k}$, and $k$ denotes the height and width of the kernel. Next, we provide detailed descriptions of spatial feature aggregation, channel feature aggregation, and weight generation.

\textbf{Spatial Feature Aggregation.} To obtain spatial-wise contextual information, the input feature is reshaped to $c \times r^2$. Then it is processed by a linear embedding layer followed by normalization and a ReLU activation. It should be noted that while this linear operation is essentially a dense feature projection, it connects every spatial position across the flattened dimension. Thus, it inherently acts as a global spatial aggregator that models the dependencies among all pixels within the patch. The result is a compact representation $G \in \mathbb{R}^{c\times m}$ that encodes the spatial information of the input features.

\textbf{Channel Feature Aggregation.} This part aims to build channel-wise feature aggregation, and it projects the input feature $G \in \mathbb{R}^{c\times m}$ to the space of output dimension $n$. This part also consists of a linear transformation, normalization, and ReLU activation. Similarly, this dense projection mathematically fuses information across the channel dimension, obtaining the channel-aggregated feature $C\in \mathbb{R}^{n \times m}$. This step effectively enables the modeling of dependencies across different channels.

\textbf{Weight Generation and Convolution.} The spatial aggregation features $G$ and channel aggregation features $C$ are jointly used to generate adaptive convolution weights. Specifically, $G$ is expanded through a linear mapping into $M'_G \in \mathbb{R}^{n \times c \times k \times k}$. $C$ is also expanded into $M'_C \in \mathbb{R}^{n \times c \times k \times k}$. Both $M'_G$ and $M'_C$ are combined to form the dynamic weight matrix $M$ as follows:
\begin{equation}
  M =  \sigma(M'_G + M'_C),
\end{equation}
where $M\in \mathbb{R}^{n \times c \times k \times k}$ has the same size as the original convolution kernel $W$, and $\sigma$ denotes the sigmoid activation function. Finally, the convolutional layer is modulated by element-wise multiplication to dynamically incorporate the global context-aware information:

\begin{equation}
  W' =  M \odot W,
\end{equation}
where $\odot$ denotes an element-wise multiplication, and $W'$ is the modulated convolution kernel. Finally, the modulated convolution kernel is applied on the input feature maps, and the contextual information is exploited to capture more representative patterns. In our following experiments, $m=4$ and $k=3$ are used.

\begin{algorithm}[t]
\caption{Global Dynamic Convolution (GDConv)}
\label{alg:gdconv}
\begin{algorithmic}[1]

\REQUIRE Input patch $I \in \mathbb{R}^{c \times r \times r}$, base kernel $W$

\ENSURE feature map $O \in \mathbb{R}^{n \times r \times r}$

 \textbf{Spatial Feature Aggregation:}
\STATE $G \leftarrow \text{Linear}(I)$
\STATE $G \leftarrow \text{Norm}(G)$
\STATE $G \leftarrow \text{ReLU}(G)$ \hfill $\triangleright G \in \mathbb{R}^{c \times m}$ ~~~~~~~~

\textbf{Channel Feature Aggregation:}
\STATE $C \leftarrow \text{Linear}(G)$
\STATE $C \leftarrow \text{Norm}(C)$
\STATE $C \leftarrow \text{ReLU}(C)$ \hfill $\triangleright C \in \mathbb{R}^{n \times m}$ ~~~~~~~~

\textbf{Weight Generation:}
\STATE $M'_G \leftarrow \text{LinearExpand}(G)$ \hfill $\triangleright M'_G \in \mathbb{R}^{n \times c \times k \times k}$
\STATE $M'_C \leftarrow \text{LinearExpand}(C)$ \hfill $\triangleright M'_C \in \mathbb{R}^{n \times c \times k \times k}$
\STATE $M \leftarrow \sigma(M'_G + M'_C)$ \hfill $\triangleright M \in \mathbb{R}^{n \times c \times k \times k}$ ~~

\textbf{Kernel Modulation and Convolution:}
\STATE $W' \leftarrow W \odot M$ \hfill $\triangleright$ Modulated kernel
\STATE $O \leftarrow \text{Conv}(I, W')$ \hfill $\triangleright O \in \mathbb{R}^{n \times r \times r}$ ~~~~~~

\RETURN $O$
\end{algorithmic}
\end{algorithm}

More details of our GDConv are provided in Algorithm \ref{alg:gdconv}. It is worth emphasizing that the proposed GDConv explicitly extracts global representations through both spatial-wise and channel-wise feature aggregations. In contrast, conventional global average pooling also captures global information, but it does so by aggressively downsampling the input feature map, which may discard discriminative details. By dynamically modeling feature dependencies in both the spatial and channel dimensions, GDConv significantly improves the expressiveness of the convolutional kernel. This enhanced representation capacity not only preserves richer contextual cues but also contributes to more accurate and robust change detection performance.

\subsection{Two-Stage Mixup Training}

By simultaneously interpolating features and labels, Mixup may produce uncertain or ambiguous training samples that do not necessarily align with realistic data distributions. Such ambiguity can introduce instability into the optimization process, often causing oscillatory behavior or slowing down the convergence of the network during training. To mitigate the inherent limitations of the standard Mixup, we propose two-stage Mixup training scheme which is more suitable for SAR change detection. The two-stage Mixup algorithm is illustrated in Algorithm \ref{al_2m}. Specifically, the total number of epochs in training is denoted as $num$, and the whole training process is divided into the following two stages:

\begin{itemize}
	
\item \textbf{Stage 1:} During the first half of training, i.e., from epoch $1$ to epoch $num/2$, the network is trained exclusively with Mixup-based data augmentation.  Here, $num/2$ is assumed to be an integer to ensure a clear division of the training process into two equal-length stages. This stage allows the model to learn more generalized representations by exposing it to interpolated samples and their corresponding soft labels.

\item \textbf{Stage 2:} In the second half of training, i.e., from epoch $num/2+1$ to epoch $num$, Mixup is applied in a probabilistic manner with probability $\varepsilon$. 
The value of $\varepsilon$ decreases linearly as training progresses, gradually reducing the frequency of Mixup augmentation. This schedule enables the model to shift from learning with ambiguous, interpolated samples toward focusing more on original, unaltered samples, thereby stabilizing optimization and improving final convergence.

\end{itemize}

In the first stage (standard Mixup), virtual feature-label vectors are generated by Mixup. In the second stage, $\varepsilon$ decreases linearly from 1 to 0. For each mini-batch, we randomly generate threshold $\theta\in[0,1]$. if $\theta<\varepsilon$, Mixup training is employed for the mini-batch. Otherwise, the training data in the mini-batch are directly used for training. For both stages, the cross-entropy loss is employed. In our implementations, 200 epochs are employed.

\begin{algorithm}[th]
\caption{Two-Stage Mixup}
\label{al_2m}
\begin{algorithmic}[1]

\REQUIRE Training dataset $(X,Y)$, number of training epochs $num$

\FOR{$i = 1$ to $num$}

    \STATE Draw a mini-batch $(x,y)$

    \IF{$i \leq num/2$}
    
        \STATE \textit{// First stage}
        \STATE $(\tilde{x},\tilde{y}) \gets \text{Mixup}(x,y)$
        
    \ELSE
    
        \STATE \textit{// Second stage}
        \STATE $\varepsilon \gets 2(num-i)/num$
        \STATE Randomly generate threshold $\theta \in [0,1]$
        
        \IF{$\theta < \varepsilon$}
            \STATE $(\tilde{x},\tilde{y}) \gets \text{Mixup}(x,y)$
        \ELSE
            \STATE $(\tilde{x},\tilde{y}) \gets (x,y)$
        \ENDIF
        
    \ENDIF

    \STATE Train model with mini-batch $(\tilde{x},\tilde{y})$

\ENDFOR

\end{algorithmic}
\end{algorithm}

As illustrated in  Algorithm \ref{al_2m}, the proposed approach enables the model to explore most of the sample representation space by applying Mixup in the first stage. In the second stage, we define an exploration rate $\varepsilon$ which denotes the probability that our model will use Mixup. As $\varepsilon$ decreases gradually, the model tends to exploit a more robust feature representation via reliable samples.

\section{Experiments and Analysis}

\subsection{Dataset and Evaluation Criteria}

We evaluated the proposed GDNet on three representative multitemporal SAR datasets. The first dataset is the Sulzberger dataset, which was acquired by the ENVISAT sensor on March 11 and March 16, 2011. It covers a region of the Sulzberger Ice Shelf in Antarctica, an area of considerable scientific interest due to its dynamic ice conditions. The other two datasets are referred to as Chao Lake I and Chao Lake II, both of which consist of SAR images of size $384 \times 384$ pixels. These datasets cover a region of Chao Lake in China and were obtained from the Sentinel-1 sensor in May 2020 and July 2020, respectively. Notably, this period corresponds to the historical highest water level recorded in Chao Lake, thereby providing challenging test cases for change detection.

To ensure reliable evaluation, the ground-truth change maps for all datasets were manually and carefully annotated by domain experts, incorporating both visual inspection and prior knowledge of the study areas. The change detection performance of GDNet was quantitatively assessed using several widely adopted metrics, including the number of false positives (FP), false negatives (FN), overall error (OE), percentage of correct classification (PCC), and the Kappa coefficient (KC). Specifically, FP represents the number of unchanged pixels that are incorrectly classified as changed. FN corresponds to the number of truly changed pixels that are mistakenly labeled as unchanged. OE measures the proportion of misclassified pixels relative to the total number of pixels, and it is computed as follows:
\begin{equation}
    \textrm{OE}= \frac{\textrm{FP} + \textrm{FN}}
    {\textrm{TP} + \textrm{TN} + \textrm{FP} + \textrm{FN}},
\end{equation}
where TP and TN denote the numbers of true positives and true negatives, respectively. A lower OE indicates better change detection performance. 

PCC reflects the proportion of correctly classified pixels, including both changed and unchanged categories. It is complementary to OE since $\textrm{PCC}=1-\textrm{OE}$. The KC measures the agreement between prediction and ground truth while accounting for agreement by chance. A KC value closer to 1 indicates more reliable and consistent change-detection results. These metrics provide a comprehensive evaluation of both the accuracy and robustness of the proposed method.

\begin{table}[h]
\centering
\renewcommand{\arraystretch}{1.4}
\caption{Ablation Studies of the Proposed GDNet on the Sulzberger Dataset}
\label{table_ablation}
\resizebox{\linewidth}{!}{
\begin{tabular}{c| c c} 
\toprule
Method & PCC ($\%$) & 
Training Time (in seconds) \\
\midrule
Basic Network & 95.34 & \textbf{84.63} \\
w/o GDConv   & 96.53  &92.60  \\  
w/o Two-Stage Mixup  & 96.64  &153.95    \\ 
\rowcolor{Bg} Proposed GDNet & \textbf{96.75}  & 159.35   \\  
\bottomrule
\end{tabular}
}
\end{table}

\subsection{Ablation Study}

In order to demonstrate the effectiveness of the proposed global dynamic convolution and two-stage Mixup, we employed three variants for comparison. First, we designed a \emph{Basic Network} with the same structure as GDNet, which uses conventional convolution instead of GDConv without Mixup. Then, we ran our model without GDConv (\emph{w/o GDConv}) and without two-stage Mixup (\emph{w/o Two-Stage Mixup}). As shown in Table \ref{table_ablation}, it is obvious that both GDConv and two-stage Mixup can boost the change detection performance (The best results are marked in bold). The training time of different model is also shown in Table \ref{table_ablation}. It can be seen that our model is improved by adding reasonable computational cost.

\begin{table}[h]
\centering
\caption{Ablation Study of the Components of GDConv on the Sulzberger Dataset}
\label{table_Components_of_GDConv}
\renewcommand{\arraystretch}{1.3}
\resizebox{\linewidth}{!}{
\begin{tabular}{l|c c c} 
\toprule
\multicolumn{1}{c|}{Modulation Branch} & PCC (\%) & Params (K) & FLOPs (K) \\ 
\midrule
Only Spatial ($G$) & 96.64 & 62.93 & 71.30 \\
Only Channel ($C$) & 96.61 & 63.06 & 72.24 \\
\rowcolor{Bg} \textbf{Complete GDConv ($G+C$)} & \textbf{96.75} & 63.22 & 73.22 \\
\bottomrule
\end{tabular}
}
\end{table}

Furthermore, to validate the necessity and effectiveness of the additive fusion between spatial and channel aggregated features within GDConv, we conducted an ablation study. Specifically, we evaluated variants using only the result of Spatial Feature Aggregation ($G$) and only Channel Feature Aggregation ($C$) for kernel modulation. As shown in Table \ref{table_Components_of_GDConv}, while the single-branch variants yield reasonable performance, the complete GDConv ($G+C$) achieves the highest PCC (96.75\%). Although $C$ is mathematically derived via a linear mapping from $G$, $G$ focuses on spatial structural contexts across the flattened dimension, whereas $C$ further distills channel-wise semantic dependencies. The results demonstrate that their additive fusion successfully captures complementary information rather than merely causing feature redundancy. More importantly, compared to the single-branch variants, the complete GDConv introduces only a negligible increase in parameters and computational cost. This strongly proves that the performance gain originates from the effective complementary fusion.

\begin{figure}
  \centering
  \includegraphics[width=\linewidth]{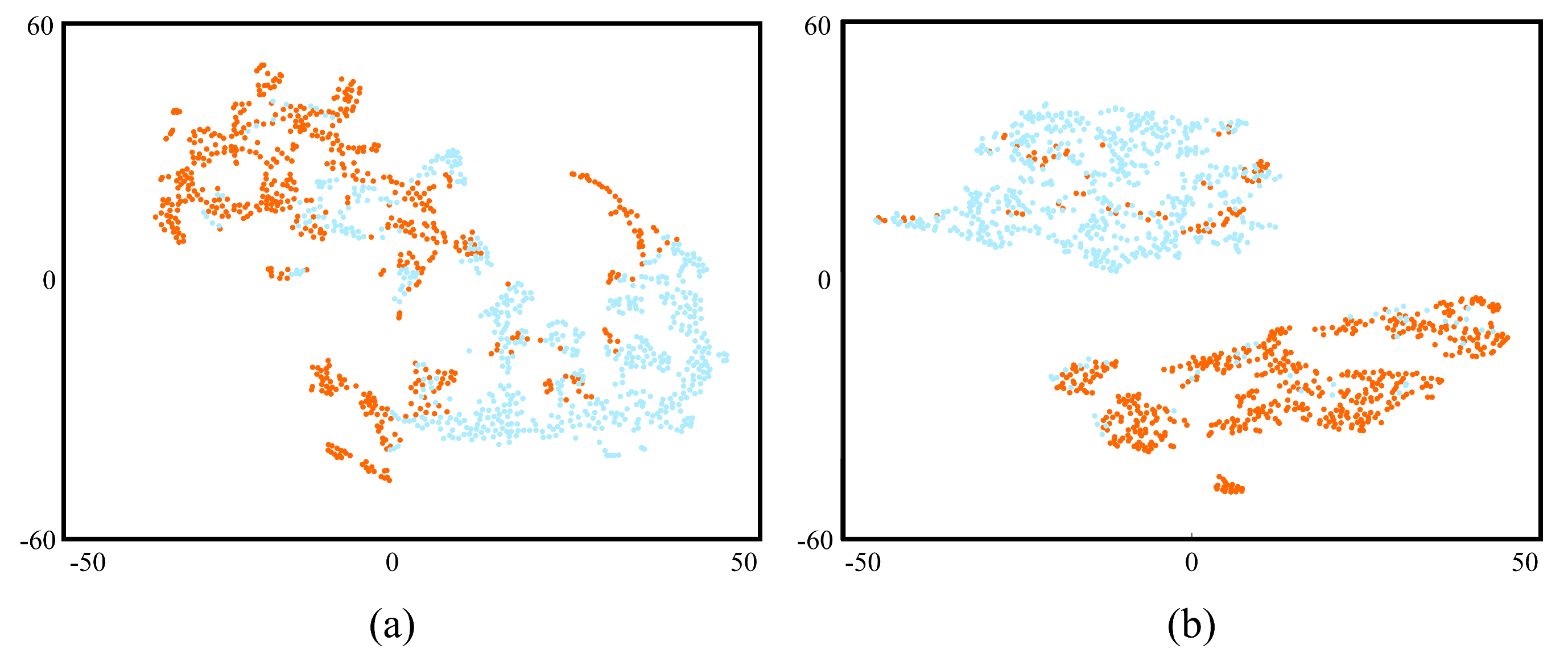}
  \caption{Visualization of the feature representations of the Sulzberger dataset. (a) Features before the GDConv. (b) Features after the GDConv.}
  \label{fig_visual}
\end{figure}

To further demonstrate the effectiveness of GDConv, we visualized the features before the last GDConv and the features after the last GDConv layer using t-SNE. It provides an intuition of how the features are arranged in a high-dimensional space. 6000 samples are selected from the testing set and each sample contains 864-dimensional feature vector. As shown in Fig. \ref{fig_visual}, the proposed GDConv successfully refines the features with more clearly defined cluster boundaries.

\subsection{Comparisons with Closely Related Attention Methods}

To evaluate the performance and advancement of the proposed module on a more comprehensive benchmark, we compared the proposed GDConv with four representative and advanced attention methods: Squeeze-and-Excitation (SE) \cite{senet}, Spatial Attention (SA) \cite{saunet}, Agent Attention (AA) \cite{agentattention}, and Progressive Focused Attention (PFA) \cite{pfa}. SE and SA are classic methods that adjust features along the channel and spatial axes, respectively. AA is a recently proposed advanced mechanism that introduces a small number of agent tokens to broadcast and aggregate information, achieving a seamless integration of highly expressive Softmax attention and efficient linear attention. PFA is another state-of-the-art method that achieves progressive feature focusing and efficient aggregation via cross-layer attention map fusion and sparse matrix multiplication.

\begin{figure}[htbp]
    \centering
    \includegraphics[width=0.9\linewidth]{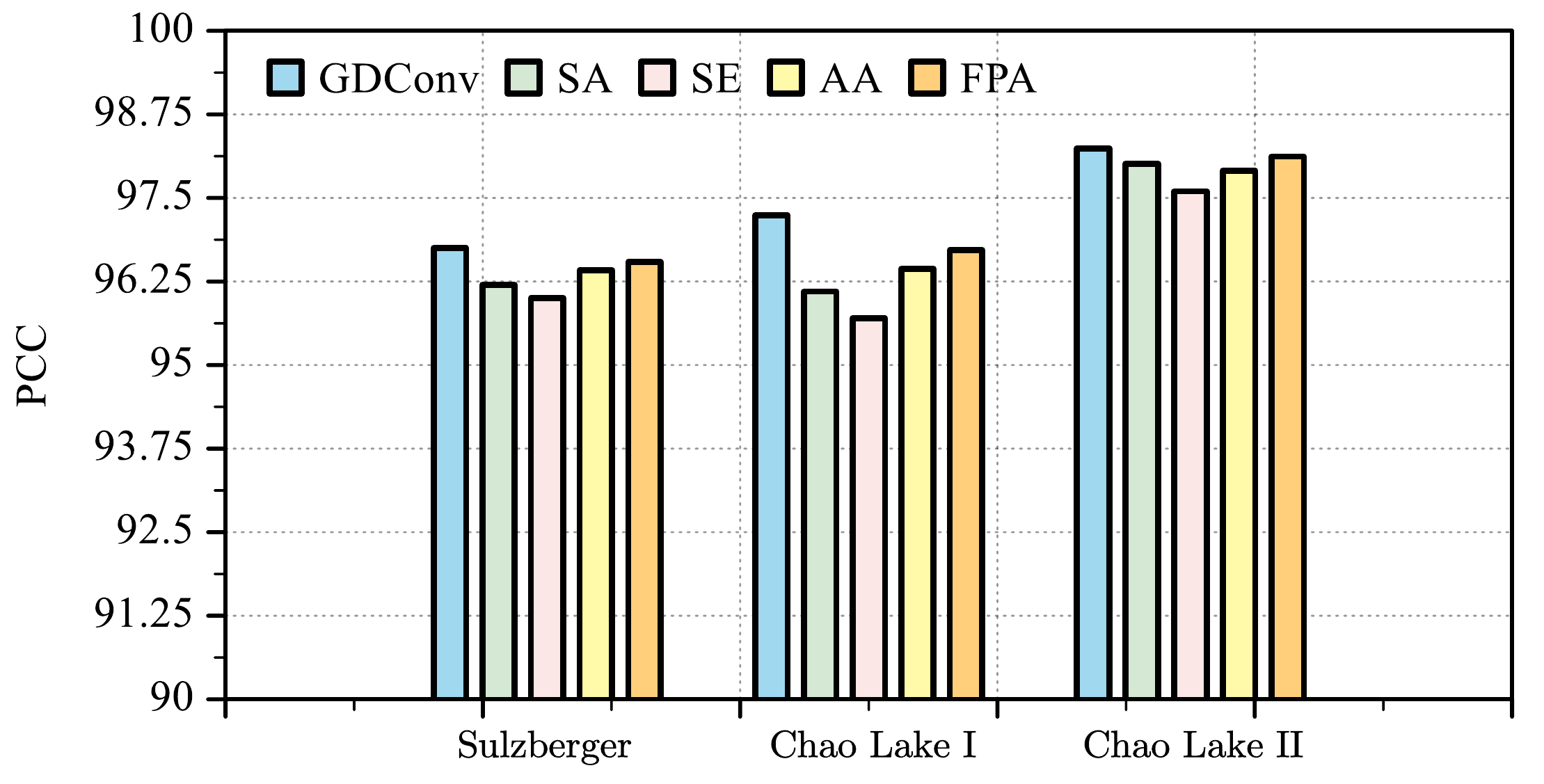}
    \caption{Experimental results of different attention methods on three datasets.}
    \label{fig_atten}
\end{figure}

Fig. \ref{fig_atten} presents the experimental results in terms of PCC of these attention methods across the three datasets. The proposed GDConv consistently outperforms all compared methods on all three datasets. This superiority demonstrates that GDConv generates more robust and discriminative features tailored for SAR change detection via dynamically modeling both spatial and channel-wise feature aggregations, effectively balancing global context modeling and local structural details preservation.

\subsection{Experimental Results and Comparison}

Nine closely related methods for change detection were compared, including PCAKM \cite{5196726}, GaborPCANet \cite{gao16grsl}, NR-ELM \cite{gao16jars}, CWNN \cite{gao19cwnn}, CAMixer \cite{zhang2023convolution}, WBANet \cite{xjw24grsl}, TSPLR \cite{fang2024unsupervised}, DSNet \cite{liu2025differential} and DBFNet \cite{dhh25grsl}.  For PCAKM, a PCA analysis is performed on the contextual information, followed by the application of the k-means algorithm to cluster the extracted features. GaborPCANet is composed of cascaded PCA layers and binary hashing layers. NR-ELM  is used to identify pixels with a high probability of change or no change, and ELM is employed to train with these pixels. CWNN is a change detection method based on Convolutional Wavelet Neural Networks, used for pixel classification. CAMixer generates training samples via hierarchical FCM and utilizes a Transformer-like architecture with PCAM and GFFN modules to jointly extract local-global features and suppress speckle noise. WBANet replaces traditional pooling with reversible wavelet downsampling on keys/values, preserving all high-frequency details while enlarging receptive fields. TSPLR relies on hierarchical FCM to separate confident and uncertain datasets for semi-supervised learning, iteratively refining pseudo-labels and employing a weighted loss function to dynamically adjust training weights. DSNet is a spatial-frequency dual-domain network that uses a difference-guided attention module to associate difference maps with original images, combining multi-scale spatial features with DCT-extracted frequency information to improve robustness. DBFNet uses dynamic shift convolution module that adaptively fuses multiple kernels and pixel offsets to overcome the representation limits of fixed CNNs. For a fair comparison, the compared methods were implemented using default parameters. To more intuitively and effectively reflect the results of various methods, visual and quantitative analyses were conducted.

\begin{figure}[htb]
  \centering
  \includegraphics[width=3.5in]{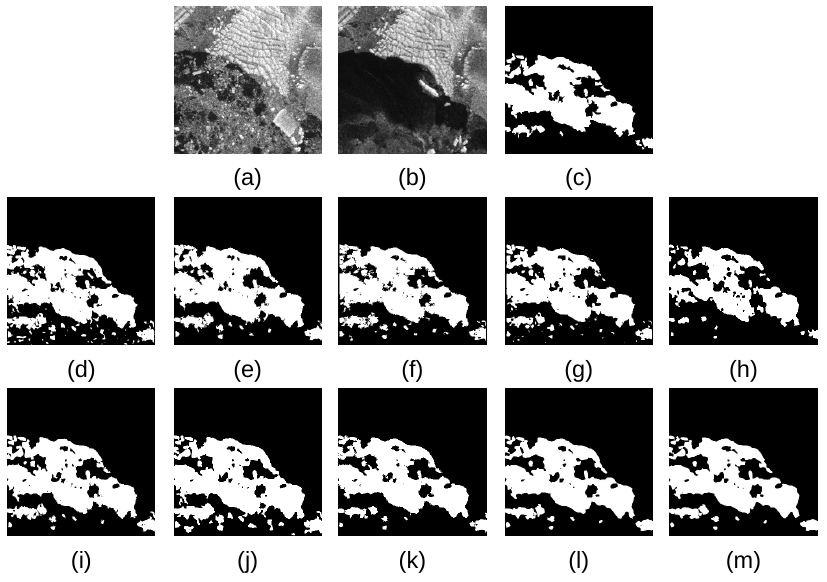}
  \caption{Visualized results of different change detection methods on the Sulzberger dataset. (a) Image acquired at T1. (b) Image acquired at T2. (c) Ground truth image. (d) Result by PCAKM. (e) Result by GaborPCANet. (f) Result by NR-ELM. (g) Result by CWNN. (h) Result by CAMixer. (i) Result by WBANet. (j) Result by TSPLR. (k) Result by DSNet. (l) Result by DBFNet. (m) Result by the proposed GDNet.}
  \label{fig_result1}
\end{figure}

\begin{table}[h]
\centering
\caption{Change Detection Results of Different Methods on the Sulzberger Dataset}
\label{table_res}
\renewcommand{\arraystretch}{1.4}
\resizebox{\linewidth}{!}{
\begin{tabular}{c|c c c c c } 
\toprule
Method
&  FP  & FN & OE (\%) & PCC (\%) & KC (\%) \\ 
\midrule
PCAKM \cite{5196726} & 3308 & 701 & 6.12 & 93.88 & 84.49 \\
GaborPCANet \cite{gao16grsl}   & 16777 & 1500 & 12.39 & 87.61 & 45.01 \\
NR-ELM \cite{gao16jars} & 2386 & 646 & 4.63 & 95.37 & 88.07\\
CWNN \cite{gao19cwnn}   & 1598 & 1132 & 4.17 & 95.83 & 88.98\\
CAMixer \cite{zhang2023convolution}   & \textbf{269} & 2362 & 4.01 & 95.99 & 86.91\\
WBANet \cite{xjw24grsl}   & 1932 & 494 & 3.70 & 96.30 & 90.40\\
TSPLR \cite{fang2024unsupervised}   & 1963 & \textbf{399} & 3.60 & 96.40 & 90.67\\
DSNet \cite{liu2025differential}   & 1554 & 711 & 3.46 & 96.54 & 90.55\\
DBFNet \cite{dhh25grsl}  & 1601 & 644 & 3.43 & 96.57 & 91.03\\
\rowcolor{Bg} Proposed GDNet  & 1398 & 731 & \textbf{3.25} & \textbf{96.75} & \textbf{91.44}\\
\bottomrule
\end{tabular}
}
\end{table}

\textit{1) Results on the Sulzberger Dataset:}
Fig. \ref{fig_result1}  presents the change detection results on the Sulzberger dataset. The corresponding quantitative metrics are shown in Table \ref{table_res}. As we can see, the result of PCAKM tends to be rather noisy, and hence it generates a rather high FP value. Furthermore, GaborPCANet, NR-ELM, CWNN, WBANet, TSPLR, DSNet and DBFNet exhibit higher FP values. Visual inspection reveals that this is due to the generation of numerous additional change regions, which impact the final results. The PCC value of the proposed GDNet is improved by 2.87$\%$, 9.14$\%$, 1.38$\%$, 0.92$\%$, 0.76$\%$, 0.45$\%$, 0.35$\%$, 0.21$\%$ and 0.18$\%$ over PCAKM, GaborPCANet, NR-ELM, CWNN, CAMixer, WBANet, TSPLR, DSNet and DBFNet, respectively. Among all the methods, the proposed GDNet produced higher PCC and KC values, and has the smallest OE value. This is due to the two-stage Mixup generating more stable sample data. In addition, the global dynamic convolution module we designed can integrate global context information into local features by interacting global information with local features. The experimental results show that our model can retain the details of changes and suppress speckle noise on the Sulzberger dataset.

\begin{figure}[ht]
  \centering
  \includegraphics[width=3.5in]{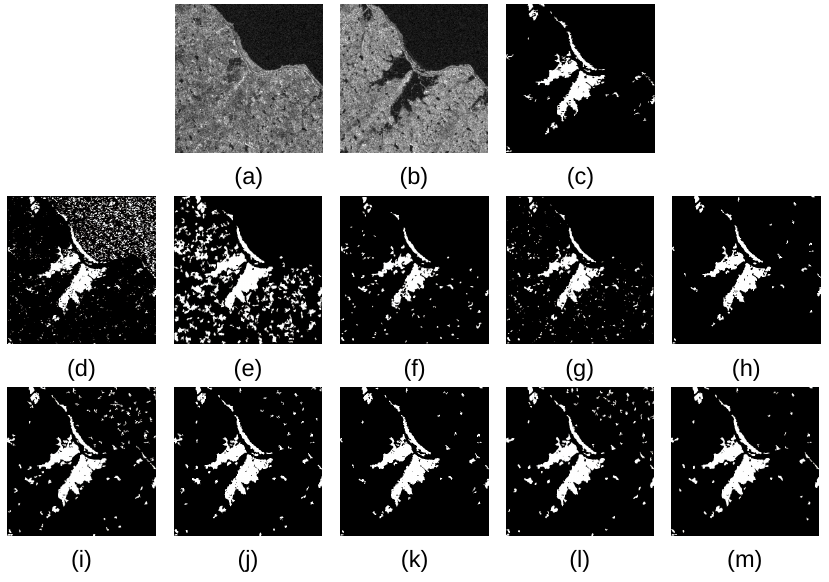}
  \caption{Visualized results of different change detection methods on the Chao Lake I dataset. (a) Image acquired at T1. (b) Image acquired at T2. (c) Ground truth image. (d) Result by PCAKM. (e) Result by GaborPCANet. (f) Result by NR-ELM. (g) Result by CWNN. (h) Result by CAMixer. (i) Result by WBANet. (j) Result by TSPLR. (k) Result by DSNet. (l) Result by DBFNet. (m) Result by the proposed GDNet.}
  \label{fig_result2}
\end{figure}

\textit{2) Results on the Chao Lake I dataset:}
Fig. \ref{fig_result2} shows the change detection results of different methods on Chao Lake I datasets, the influence of speckle noise is much stronger. It represents a more complicated situation to obtain the changed details of this flood disaster. Table \ref{table_res2} illustrates the corresponding quantitative results of different methods. As shown in Table \ref{table_res2}, we can see that in addition to NR-ELM, CWNN and the proposed GDNet, other methods have high FP values, which indicates some unchanged areas are falsely detected as changed ones. Visual inspection reveals that most methods are significantly impacted by speckle noise. However, the proposed GDNet produces a highly accurate and reliable change map, resulting in the smallest FN value. Furthermore, Table \ref{table_res2} demonstrates that GDNet yields the best PCC value of 97.24\%, achieving an improvement of at least 0.67\% compared to the other compared methods. Evidently, by optimally utilizing global features, GDNet can successfully mitigate the effect of speckle noise present in the training samples. These experimental results indicate that GDNet exhibits strong noise suppression capabilities on the Chao Lake I dataset.

\renewcommand{\arraystretch}{1.4}
\begin{table}[h]
\centering
\caption{Change Detection Results of Different Methods on Chao Lake I Datasets}
\label{table_res2}
\resizebox{\linewidth}{!}{
\begin{tabular}{c|c c c c c } 
\toprule
Method 
& FP & FN & OE (\%) & PCC (\%)  & KC (\%) \\ 
\midrule
PCAKM \cite{5196726} & 13199 & 2996 & 10.98 & 89.02 & 43.99 \\
GaborPCANet \cite{gao16grsl}   & 3987 & 1724 & 3.87 & 96.13 & 71.59 \\
NR-ELM \cite{gao16jars} & \textbf{2252} & 3552 & 3.94 & 96.06 & 69.63\\
CWNN \cite{gao19cwnn}   & 2908 & 2808 & 3.88 & 96.12 & 71.85\\
CAMixer \cite{zhang2023convolution}   & 3718 & 2079 & 3.93 & 96.07 & 70.45\\
WBANet \cite{xjw24grsl} & 4983 & 1217 & 4.20 & 95.80 & 73.54\\
TSPLR \cite{fang2024unsupervised}   & 4534 & 1052 & 3.79 & 96.21 & 74.04\\
DSNet \cite{liu2025differential}   & 3917 & 1147 & 3.43 & 96.57 & 74.91\\
DBFNet \cite{dhh25grsl}  & 4172 & 1170 & 3.62 & 96.38 & 76.55\\
\rowcolor{Bg} Proposed GDNet  & 3029 & \textbf{1034} & \textbf{2.76} & \textbf{97.24} & \textbf{81.46}\\
    \bottomrule
    \end{tabular}
}
\end{table}

\begin{figure}[ht]
  \centering
  \includegraphics[width=3.5in]{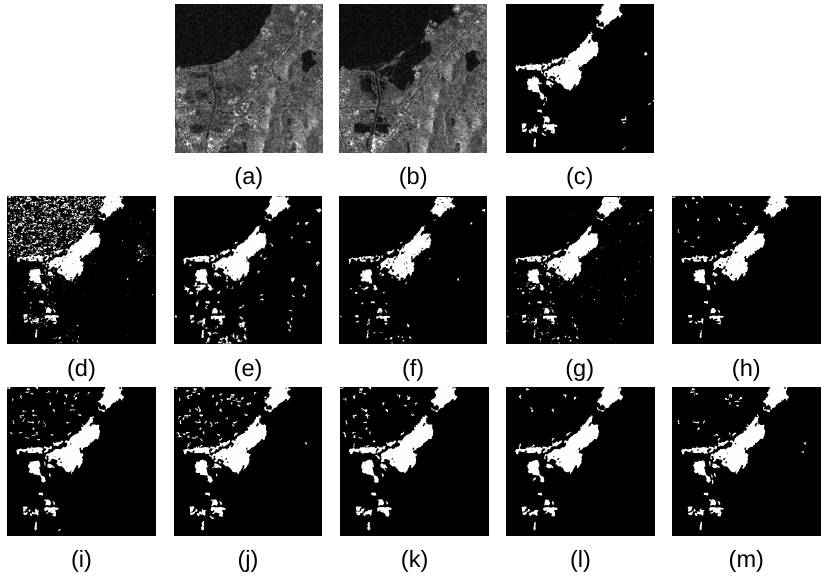}
  \caption{Visualized results of different change detection methods on the Chao Lake II dataset. (a) Image acquired at T1. (b) Image acquired at T2. (c) Ground truth image. (d) Result by PCAKM. (e) Result by GaborPCANet. (f) Result by NR-ELM. (g) Result by CWNN. (h) Result by CAMixer. (i) Result by WBANet. (j) Result by TSPLR. (k) Result by DSNet. (l) Result by DBFNet. (m) Result by the proposed GDNet.}
  \label{fig_result3}
\end{figure}

\textit{3) Results on the Chao Lake II dataset:}
Fig. \ref{fig_result3} presents the change detection results on the Chao Lake II dataset. The corresponding quantitative metrics are listed in Table \ref{table_res3}. The results of NR-ELM, and CWNN miss many changed regions, thus these methods suffer from very high FN values. For PCAKM, GaborPCANet, CAMixer, WBANet, TSPLR and DSNet, the values of FP are relatively high, which can be partly attributed to noisy samples. In contrast, the proposed GDNet generates a change map that is much closer to the ground truth. The detected change regions are well aligned with the reference, with reduced noise, sharper boundaries, and better completeness. These results highlight the advantage of incorporating global dynamic convolution and two-stage Mixup in improving both the accuracy and robustness of SAR image change detection, particularly in complex environments such as the Chao Lake II scene.

\begin{table}[h]
\centering
\caption{Change Detection Results of Different Methods on Chao Lake II Datasets}
\label{table_res3}
\renewcommand{\arraystretch}{1.4}
\resizebox{\linewidth}{!}{
\begin{tabular}{c|c c c c c } 
\toprule
Method 
& FP & FN & OE (\%) & PCC (\%)  & KC (\%) \\ 
\midrule
PCAKM \cite{5196726} & 8521 & 2248 & 7.30 & 92.70 & 65.58 \\
GaborPCANet \cite{gao16grsl}   & 2946 & 1771 & 3.20 & 96.80 & 82.66 \\
NR-ELM \cite{gao16jars} & \textbf{595} & 3836 & 3.00 & 97.00 & 81.27\\
CWNN \cite{gao19cwnn}   & 959 & 2397 & 2.28 & 97.72 & 86.63\\
CAMixer \cite{zhang2023convolution}   & 2453 & 1533 & 2.70 & 97.30 & 84.87\\
WBANet \cite{xjw24grsl}   & 3107 & 779 & 2.64 & 97.36 & 86.18\\
TSPLR \cite{fang2024unsupervised}   & 2641 & \textbf{651} & 2.23 & 97.77 & 88.18\\
DSNet \cite{liu2025differential}   & 2350 & 720 & 2.08 & 97.92 & 87.62\\
DBFNet \cite{dhh25grsl}  & 1350 & 1334 & 1.82 & 98.18 & 89.78\\
\rowcolor{Bg} Proposed GDNet  & 1821 & 768 & \textbf{1.76} & \textbf{98.24} & \textbf{90.56}\\
    \bottomrule
    \end{tabular}
}
\end{table}

\textit{4) Efficiency and Model Complexity Analysis:}
To systematically evaluate the computational efficiency and model complexity of the proposed framework, we conducted a comparative analysis of floating-point operations (FLOPs) and parameter counts. Table \ref{table_efficiency} delineates the efficiency metrics of various deep learning-based approaches evaluated on the Sulzberger dataset. As demonstrated, the proposed GDNet achieves a favorable trade-off between detection accuracy and computational cost. Specifically, GDNet registers a computational footprint of 73.22K FLOPs, which is markedly lower than those of recent state-of-the-art architectures, such as CAMixer, TSPLR, and the computationally intensive DSNet. Furthermore, the parameter capacity of GDNet is constrained to 63.22K, ensuring a relatively lightweight profile compared to highly parameterized models like DSNet. Although WBANet presents lower complexity metrics, it incurs a performance degradation of 0.45\% in PCC relative to our framework. Consequently, these quantitative findings substantiate that the superior performance of GDNet inherently stems from the structural efficacy of the global dynamic context-aware mechanism.

\begin{table}[h]
\centering
\caption{Efficiency Comparison of Different Methods on the Sulzberger Dataset}
\label{table_efficiency}
\renewcommand{\arraystretch}{1.4}
\begin{tabular}{c|c c c} 
\toprule
Method & Parameter(K) & FLOPs(K) & PCC(\%)  \\ 
\midrule
GaborPCANet \cite{gao16grsl}   & 9.10 & 735.50 & 87.61\\
CWNN \cite{gao19cwnn}   & 20.69 & 221.28 & 95.83\\
CAMixer \cite{zhang2023convolution}   & 34.21 & 287.99 & 95.99\\
WBANet \cite{xjw24grsl}   & 7.16 & 22.59 & 96.30\\
TSPLR \cite{fang2024unsupervised}   & 57.75 & 160.20 & 96.40\\
DSNet \cite{liu2025differential}   & 129.97 & 2191.82 & 96.54\\
DBFNet \cite{dhh25grsl}  & 11.50 & 219.09 & 96.57\\
\rowcolor{Bg} Proposed GDNet  & 63.22 & 73.22 & \textbf{96.75}\\
    \bottomrule
    \end{tabular}
\end{table}

\subsection{Discussion}

The experimental results across multiple challenging SAR datasets substantiate the superiority of the proposed GDNet. The fundamental driver of this performance lies in the synergistic effect of the Global Dynamic Convolution (GDConv) and the two-stage Mixup strategy. Unlike conventional static CNNs that rely on content-agnostic kernels or Transformer architectures that incur heavy computational overhead, GDConv adaptively modulates local convolution kernels using global spatial and channel-wise aggregated representations. This unique mechanism enables the network to actively emphasize subtle change-related features while inherently suppressing the severe speckle noise characteristic of SAR imagery. Furthermore, the two-stage Mixup training effectively bridges the gap between data diversity and optimization stability. By linearly decaying the augmentation probability in the later stages, the model successfully avoids the label ambiguity issues often caused by aggressive data augmentation, thereby ensuring a robust convergence even with limited pseudo-labeled samples.

Nevertheless, we also recognize the limitations of the current framework. The proposed method still relies on a hierarchical fuzzy c-means (FCM) clustering strategy to generate pseudo-labels during the preclassification stage. While this strategy yields highly reliable samples for self-supervised training, it introduces a multi-step pipeline that restricts the ability to fully exploit the enormous and continuously growing volume of remote sensing data. As part of our future work, we aim to investigate purely end-to-end self-supervised change detection frameworks—such as masked image modeling or contrastive learning paradigms—that eliminate the dependence on heuristic preclassification, thereby enabling more efficient, scalable, and fully automated analysis of massive SAR archives.

\section{Conclusions}

In this paper, we proposed a novel Global Dynamic Context-Aware Network (GDNet) tailored for self-supervised SAR image change detection. We introduced a dynamic convolution module that adaptively adjusts kernel weights in response to the global semantic information extracted from the input patch, successfully balancing long-range dependency modeling and fine-grained local detail preservation. Additionally, we designed a two-stage Mixup strategy to stabilize model convergence under limited and noisy data conditions. Comprehensive experiments conducted on three benchmark SAR datasets confirm the superiority of GDNet over a wide range of state-of-the-art methods. The proposed framework consistently produces highly accurate change maps with robust noise suppression and precise boundary delineation, demonstrating strong potential for real-world remote sensing change monitoring applications.

\bibliographystyle{IEEEtran}
\bibliography{ref} 

\end{document}